\documentclass[twoside,11pt]{article}

\usepackage{jmlr2e}
\usepackage{amsmath}
\usepackage{url}
\usepackage{subcaption}

\ShortHeadings{torch.manual\_seed(3407) is all you need}{D. Picard}
\firstpageno{1}

\begin{document}

\title{\texttt{torch.manual\_seed(3407)} is all you need: On the influence of random seeds in deep learning architectures for computer vision}

\author{\name David Picard \email david.picard@enpc.fr \\
       \addr LIGM, {É}cole des Ponts, 77455 Marnes la vallée, France}

\editor{}

\maketitle

\begin{abstract}
In this paper I investigate the effect of random seed selection on the accuracy when using popular deep learning architectures for computer vision. I scan a large amount of seeds (up to $10^4$) on CIFAR 10 and I also scan fewer seeds on Imagenet using pre-trained models to investigate large scale datasets. The conclusions are that even if the variance is not very large, it is surprisingly easy to find an outlier that performs much better or much worse than the average.
\end{abstract}


\section{Introduction}

In this report, I want to test the influence of the random generator seed on the results of ubiquitous deep learning models used in computer vision.
I ask the following question:
\begin{enumerate}
    \item What is the distribution of scores with respect to the choice of seed?
    \item Are there \emph{black swans}, \textit{i.e.}, seeds that produce radically different results?
    \item Does pretraining on larger datasets mitigates variability induced by the choice of seed?
\end{enumerate}
I feel these questions are important to ask and test because with the rise of large training sets and large deep learning models, it has become common practice in computer vision (and to other domains relying heavily on machine learning) to report only a single run of the experiment, without testing for its statistical validity. This trend started when computing power was limited, and it is perfectly understandable that the result of a single experiment is better than no result at all. To the best of my understanding, this seems to happen all the time in physics - first results get quickly announced,  and are later confirmed or denied. The deep learning community has what may be a faulty approach in that it never confirms results by running as many reproductions as required. Just as the noise in the measurement of a physical experiment, we know there are random factors in a deep learning experiment, like the split between training and test data, the random initialization, the stochasticity of the optimization process, etc. It has been shown in \cite{mehrer2020individual} that these affect the obtained networks in such a way that they differ measurably. I strongly believe it is important to have a sense of the influence of these factors on the results in terms of evaluation metrics. This is essentially what was done by \cite{MLSYS2021_cfecdb27} on different tasks, including computer vision, but limited to smaller datasets and \emph{only} 200 runs. Instead of performing the tedious work of measuring the influence of each source of randomness at larger scales, I propose instead to ask the much simpler 3 questions above, and that are largely based on the choice of a random seed. The variations I measure are thus the accumulation of all random factors relying on the random seed and if they are small enough, then the corresponding random factors have negligible influence. On the contrary, if they are large enough to not be negligible, then I believe deep learning publications should start to include a detailed analysis of the contribution of each random factor to the variation in observed performances.

The remaining of this report is organized as follows: First, I detail the experimental setup used and motivate my choice. Then I discuss the limitations of this experiments and how they affect the conclusions we can make from it. After that, I show the findings with respect to each question with the chosen experimental setup. Finally, I draw some conclusions from these findings bounded by the limitations exposed earlier.

\section{Experimental setup}

Because this is a simple experiment without the need to produce the best possible results, I allowed myself a budget of 1000 hours of V100 GPU computing power. This budget is nowhere near what is currently used for most of major computer vision publications, but it is nonetheless much more than what people that do not have access to a scientific cluster can ever do. All the architectures and the training procedures have thus been optimize to fit this budget constraint. 

This time was divided into half for finding the architectures and training parameters that would allow a good trade-off between accuracy and training time; and half to do the actual training and evaluation. The time taken to compute statistics and produce figures was of course not taken into account. To ensure reproducible fondings, all codes and results are publicly available\footnote{\url{https://github.com/davidpicard/deepseed}}.

\subsection{CIFAR 10}
For the experiments on CIFAR 10, I wanted to test a large number of seeds, and so training had to be fast. I settled on an operating point of 10 000 seeds, each taking 30 seconds to train and evaluate. The total training time is thus 5000 minutes or about 83 hours of V100 computing power. The architecture is a custom ResNet with 9 layers that was taken from the major results of the DAWN benchmark, see \cite{coleman2019analysis}, with the following layout:

\begin{align*}
    C^3_{64}-C^3_{128}-M^2-R[C^3_{128}-C^3_{128}]-C^3_{256}-M^2-C^3_{256}-M^2-R[C^3_{256}-C^3_{256}]-G^1-L_{10}
\end{align*}

where $C^k_d$ is a convolution of kernel size $k$ and $d$ output dimension followed by a batch normalization and a ReLU activation, $M^2$ is a max pooling of stride 2, $R[\cdot]$ denote a skip connection of what is inside the brackets, $G^1$ is a global max pooling and $L_{10}$ is a linear layer with 10 output dimensions.

The training was performed using a simple SGD with momentum and weight decay. The loss was a combination of a cross-entropy loss with label-smoothing and the regular categorical cross-entropy. The learning rate scheduling was a short increasing linear ramp followed by a longer decreasing linear ramp.

To make sure that the experiment on CIFAR 10 was close to convergence, I ran a longer training setup of 1 minute on only 500 seeds, for a total of just over 8 hours.
The total training time for CIFAR was thus just over 90 hours of V100 computing time.

\subsection{ImageNet}
For the large scale experiments on ImageNet, it was of course impossible to train commonly used neural networks from scratch in a reasonable amount of time to allow gathering statistics over several seeds. I settled on using pretrained networks where only the classification layer is initialized from scratch. I used three different architectures/initializations: 
\begin{itemize}
    \item Supervised ResNet50: the standard pretrained model from pytorch;
    \item SSL ResNet50: a ResNet50 pretrained with self-supervision from the DINO repository, see \cite{caron2021emerging};
    \item SSL ViT: a Visual transformer pretrained with self-supervision from the DINO repository, see \cite{caron2021emerging}.
\end{itemize}

All training were done using a simple SGD with momentum and weight decay optimization, with a cosine annealing scheduling that was optimized to reach decent accuracy in the shortest amount of time compatible with the budget. The linear classifier is trained for one epoch, and the remaining time is used for fine-tuning the entire network.

All model were tested on 50 seeds. The supervised pretrained ResNet50 had a training time of about 2 hours, with a total training time of about 100 hours. The SSL pretrained ResNet50 had a training time of a little over 3 hours for a total of about 155 hours of V100 computation time. This longer time is explained by the fact that it required more steps to reach comparable accuracy when starting from the SLL initialization. The SSL pretrained ViT took about 3 hours and 40 minutes per run, for a total of about 185 hours of V100 computation time. This longer training time is also explained by the higher number of steps required, but also by the lower throughput of this architecture.

The total time taken on ImageNet was about 440 hours of V100 computing time.

\section{Limitations}
\label{sec:limiations}

There are several limitations to this work that affect the conclusion that can be drawn from it. These are acknowledged and tentatively addressed here.

First, the accuracy obtained in these experiments are not at the level of the state of the art. This is because of the budget constraint that forbids training for longer times. One could argue that the variations observed in these experiments could very well disappear after longer training times and/or with the better setup required to reach higher accuracy.
This is a clear limitation of my experiments - caused by necessity - and I can do very little to argue against it. Models are evaluated at convergence, as I show in the next section. On the accuracy on CIFAR 10, the long training setup achieves an average accuracy of $90.7\%$ with the maximum accuracy being $91.4\%$, which is very well below the current state of the art. However, recall that the original ResNet obtained only $91.25\%$ with a 20 layers architecture. Thus, my setup is roughly on par with the state of the art of 2016. This means papers from that era may have been subject to the sensitivity to the seed that I observe.
Also, Note that the DAWN benchmark required to submit 50 runs, at least half of which had to perform over $94\%$ accuracy to validate the entry. In these test a good entry\footnote{\url{https://github.com/apple/ml-cifar-10-faster/blob/master/run_log.txt}} can achieve a minimum accuracy of $93.7\%$, a maximum accuracy of $94.5\%$ while validating 41 runs over $94\%$ accuracy. Thus, even with higher accuracy, there is still a fair amount of variation.
Although it seems intuitive that the variation lessens with improved accuracy, it could also be that the variations are due to the very large number of seeds tested.

For ImageNet, the situation is different, because the baseline ResNet50 obtains an average accuracy of $75.5\%$, which is not very far from the original model at $76.1\%$. The results using SSL are also close to that of the original DINO paper, in which the authors already acknowledge that tweaking hyperparameters produces significant variation.
If anything, I would even argue that the experiments on ImageNet underestimate the variability because they all start from the same pretrained model, which means that the effect of the seed is limited to the initialization of the classification layer and the optimization process.

Still, this work has serious limitations that could be overcome by spending ten times the computation budget on CIFAR to ensure all models are trained to the best possible, and probably around 50 to 100 times the computation budget on ImageNet to allow training models from scratch.
findings may thus be limited in that the observed variation is likely to be reduced when considering more accurate architectures and training setup, although I do not expect it to disappear entirely.

\section{Findings}

The findings of my experiments are divided into three parts. First, I examine the long training setup on CIFAR 10 to evaluate the variability after convergence. This should answer question 1. Then, I evaluate the short training setup on CIFAR 10 on 10000 seeds to answer question 2. Finally, I investigate training on ImageNet to answer question 3.

\subsection{Convergence instability}

\begin{figure}
    \centering
\begin{minipage}[t]{0.47\textwidth}
    \includegraphics[width=1\textwidth]{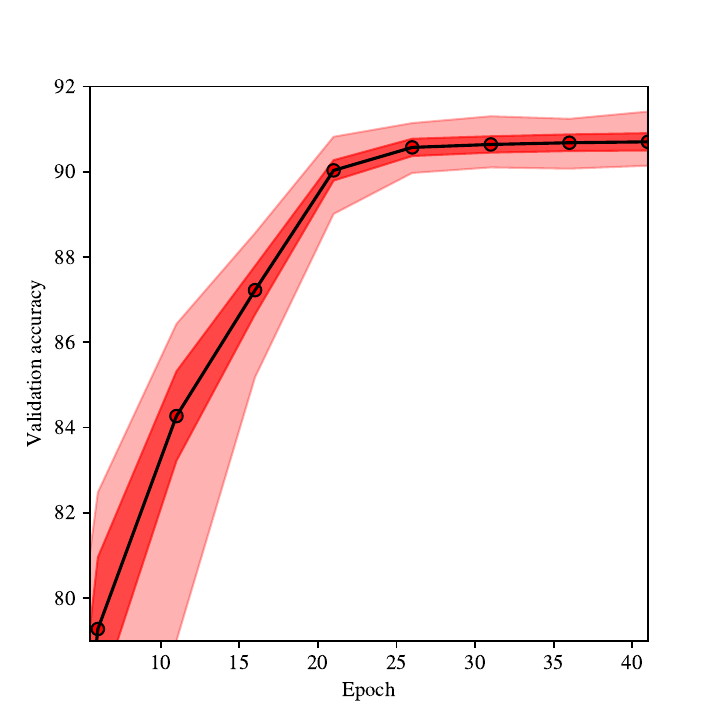}
    \caption{Validation accuracy variation on CIFAR 10 for the Resnet9 architecture against number of training epochs. he solid line represents the mean over 500 seeds, the dark red area corresponds to one standard deviation and the light red corresponds to the maximum and minimum values.}
    \label{fig:c10l_500}
\end{minipage}
\hfill
\begin{minipage}[t]{0.47\textwidth}
    \includegraphics[width=1\textwidth]{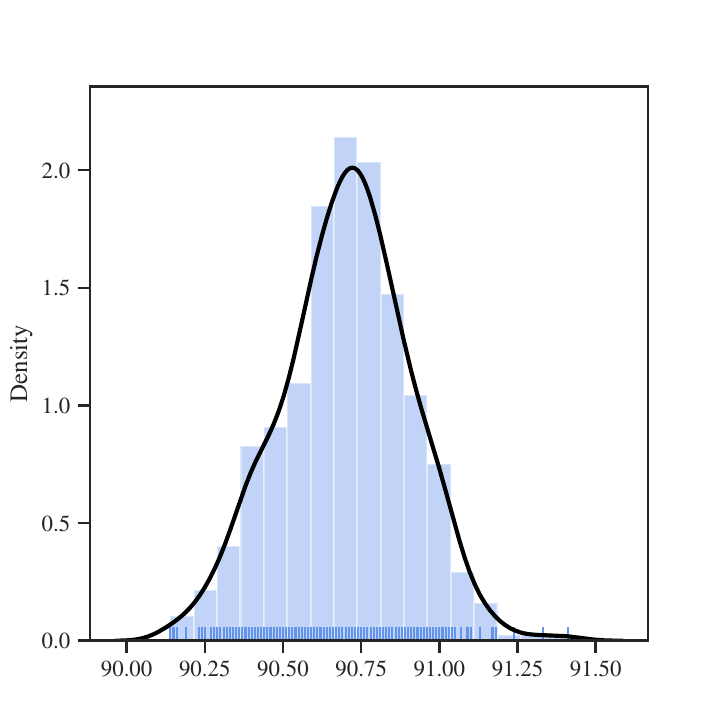}
    \caption{Histogram and density plot of the final validation accuracy on CIFAR 10 for the Resnet9 architecture over 500 seeds. Each dash at the bottom corresponds to one run.}
    \label{fig:c10l_500_d}
\end{minipage}
\end{figure}

Training on CIFAR 10 for 500 different seeds produces an evolution of the validation accuracy against time as shown on \autoref{fig:c10l_500}.
In this figure, the black line is the average, the dark red corresponds to one standard deviation and the light red corresponds to the minimum and maximum values attained across seeds.
As we can see, the accuracy does not increase past epoch 25, which means that the optimization converged.
However, there is no reduction in variation past convergence which indicates that training for longer is not likely to reduce the variation.
If it were, we should observe a narrowing light red area, whereas it remains constant. Note that this narrowing behavior is observed before convergence.

I next show the distribution of accuracy across seeds on \autoref{fig:c10l_500_d}. In this figure, I show the histogram (blue bars) with a kernel density estimation (black line) and each run (small stems at the bottom). Corresponding statistics are in \autoref{tab:c10l}.
As we can see, the distribution is mono-modal and fairly concentrated (dare I even say that it looks leptokurtic without going the trouble of computing higher order moments).
Nonetheless, we can clearly see that results around 90.5\% accuracy are numerous and as common as results around 91.0\% accuracy.
Hence a 0.5\% difference in accuracy could be entirely explained by just a seed difference, without having chosen a particularly \emph{bad} or \emph{good} seed.

The answer to first question is thus that the distribution of scores with respect to the seeds is fairly concentrated and sort of pointy. This is a reassuring result because it means that scores are likely to be representative of what the model and the training setup can do, expect when one is actively searching (knowingly or unknowingly) for a good/bad seed.

\begin{table}[h]
    \centering
    \begin{tabular}{l|c|c|c}
        Training mode & Accuracy mean $\pm$ std & Minimum accuracy & Maximum accuracy   \\
        \hline
         long & 90.70 $\pm$ 0.20 & 90.14 &  91.41\\
         short & 90.02 $\pm$ 0.23 & 89.01 & 90.83
    \end{tabular}
    \caption{Results on cifar 10 for the long and the short training setups.}
    \label{tab:c10l}
\end{table}

\subsection{Searching for \emph{Black Swans}}

I next use the short training setup to scan the first $10^4$ seeds.
The accuracy is reported in the second line of \autoref{tab:c10l}, and we can see that the average is slightly less than with the long training at around $90\%$. The interesting values of this test are the minimum and maximum accuracy among all run (obtained at the end of training), which is this case go from $89.01\%$ to $90.83\%$, that is, a $1.82\%$ difference. Such difference is widely considered as significant in the community - to the point of being an argument for publication at very selective venues - whereas we know here that this is just the effect of finding a lucky/cursed seed.

\begin{figure}
    \centering
\begin{minipage}[t]{0.47\textwidth}
    \includegraphics[width=1\textwidth]{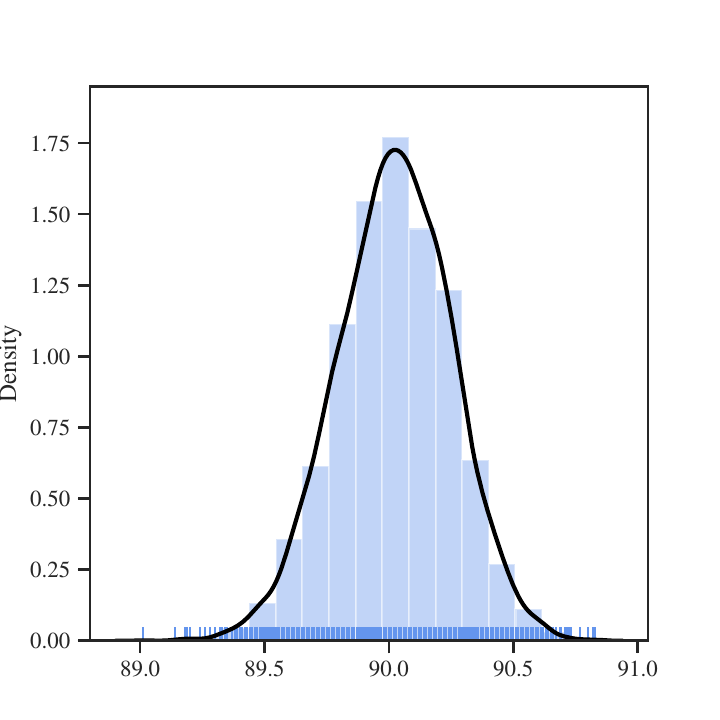}
    \caption{Histogram and density plot of the final validation accuracy on CIFAR 10 for the Resnet9 architecture over $10^4$ seeds. Each dash at the bottom corresponds to one run.}
    \label{fig:c10_10k}
\end{minipage}
\hfill
\begin{minipage}[t]{0.47\textwidth}
    \includegraphics[width=1\textwidth]{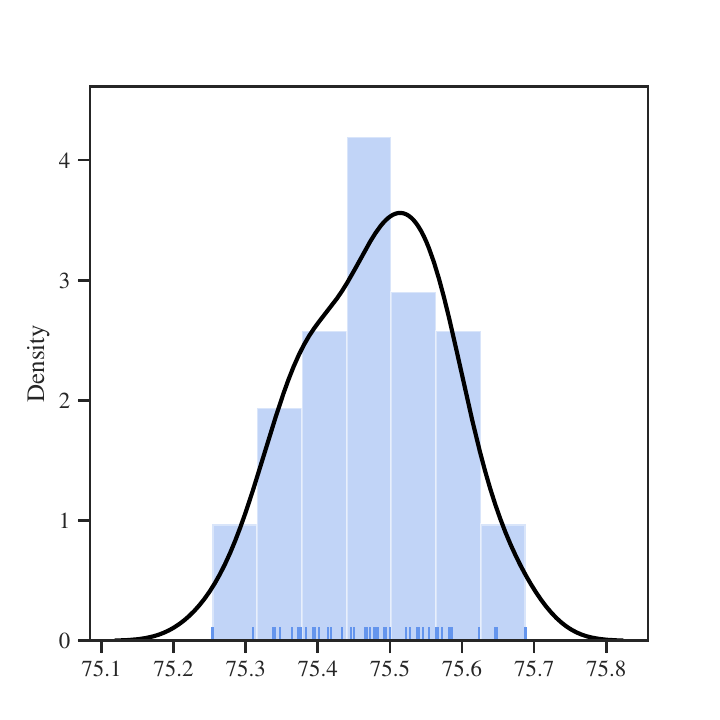}
    \caption{Histogram and density plot of the final validation accuracy on Imagenet for a pretrained ResNet50. Each dash at the bottom corresponds to one run.}
    \label{fig:im}
\end{minipage}
\end{figure}

The distribution can be seen on \autoref{fig:c10_10k} and is well concentrated between $89.5\%$ and $90.5\%$. It seems unlikely that higher or lower extremes could be obtained without scanning a very large amount of seeds. That being said, the extremes obtained by scanning only the first $10^4$ seeds are highly non-representative of what the model would usually do.

The results of this test allow me to answer positively to the second question: there are indeed seeds that produce scores sufficiently good (respectively bad) to be considered as a significant improvement (respectively downgrade) by the computer vision community. This is a worrying result as the community is currently very much score driven, and yet these can just be artifacts of randomness.

\subsection{Large scale datasets}

Next, I use the large scale setup with pretrained models fine-tuned and evaluated on Imagenet to see if using a larger training set in conjunction with a pretrained model does mitigate the randomness of the scores induced by the choice of the seed. 

The accuracies are reported in \autoref{tab:im}. Standard deviations are around $0.1\%$ and the gap between minimum and maximum values is only about $0.5\%$. This is much smaller than for the CIFAR 10 tests, but is nonetheless surprisingly high given that: 1) all runs share the same initial weights resulting from the pretraining stage except for the last layer ; 2) only the composition of batches varies due to the random seed. A $0.5\%$ difference in accuracy on ImageNet is widely considered significant in the computer vision community - to the point of being an argument for publication at very selective venues - whereas we know here that it is entirely due to the change of seed.

\begin{table}[h]
    \centering
    \begin{tabular}{l|c|c|c}
        Training mode & Accuracy mean $\pm$ std & Minimum accuracy & Maximum accuracy   \\
        \hline
         ResNet50 & 75.48 $\pm$ 0.10 & 75.25 &  75.69\\
         ResNet50 SSL & 75.15 $\pm$ 0.08 & 74.98 & 75.35\\
         ViT SSL & 76.83 $\pm$ 0.11 & 76.63 & 77.09
    \end{tabular}
    \caption{Results on Imagenet for different architecture and pretraining schemes.}
    \label{tab:im}
\end{table}

\begin{figure}
    \centering
\begin{minipage}[t]{0.47\textwidth}
    \includegraphics[width=1\textwidth]{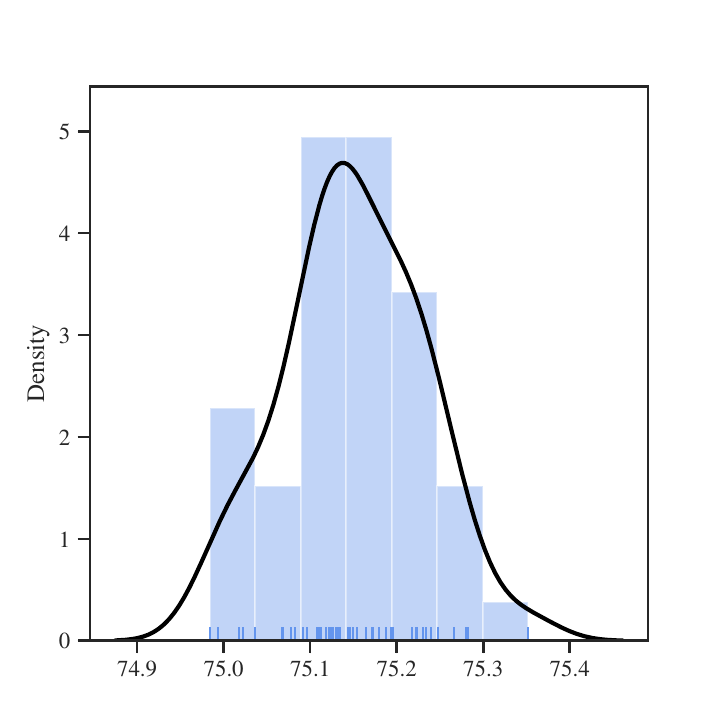}
    \caption{Histogram and density plot of the final validation accuracy on Imagenet for a self-supervised pretrained ResNet50. Each dash at the bottom corresponds to one run.}
    \label{fig:imssl}
\end{minipage}
\hfill
\begin{minipage}[t]{0.47\textwidth}
    \includegraphics[width=1\textwidth]{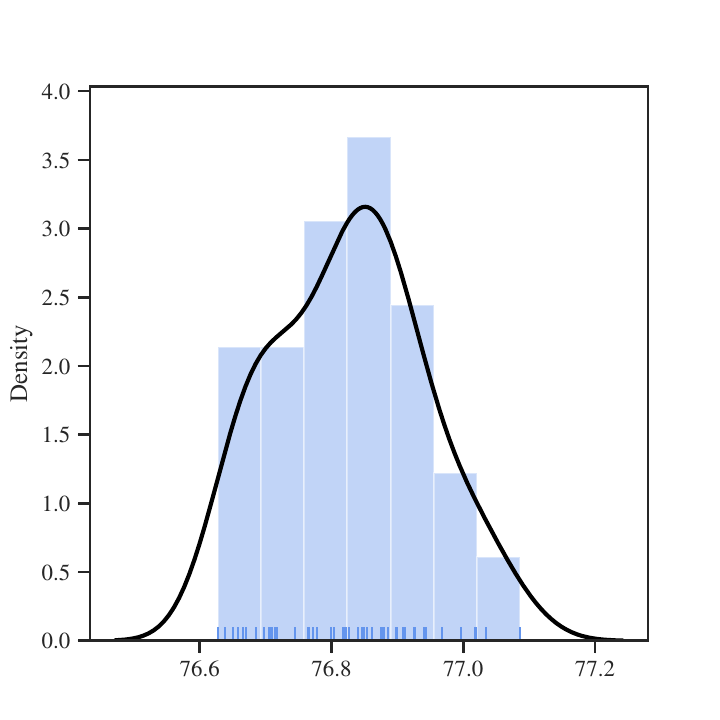}
    \caption{Histogram and density plot of the final validation accuracy on Imagenet for a self-supervised pretrained Visual Transformer. Each dash at the bottom corresponds to one run.}
    \label{fig:imsslvit}
\end{minipage}
\end{figure}

 The corresponding distributions are shown on \autoref{fig:im}, \autoref{fig:imssl} and \autoref{fig:imsslvit}. These are not as well defined as the ones obtained on CIFAR 10, and I attribute it to only using 50 seeds. In particular, looking at these distributions does not inspire me confidence that the gap would have been limited to less than $1\%$ had I scanned more than just 50 seeds.

The answer to the third question is thus mixed: In some sense, yes using pretrained models and larger training sets reduces the variation induced by the choice of seed. But that variation is still significant with respect to what is considered an improvement by the computer vision community. This is a worrying result, especially since pretrained models are largely used. If even using the same pretrained model fine-tuned on a large scale dataset leads to significant variations by scanning only 50 seeds, then my confidence in the robustness of recent results when varying the initialization (different pretrained model) and scanning a large number of seeds is very much undermined.

\section{Discussion}

I first summarize the findings of this small experiment with respect to the three opening questions: 

\paragraph{What is the distribution of scores with respect to the choice of seed?}
The distribution of accuracy when varying seeds is relatively pointy, which means that results are fairly concentrated around the mean. Once the model converged, this distribution is relatively stable which means that some seed are intrinsically better than others.

\paragraph{Are there \emph{black swans}, \textit{i.e.}, seeds that produce radically different results?}
Yes. On a scanning of $10^4$ seeds, we obtained a difference between the maximum and minimum accuracy close to $2\%$ which is above the threshold commonly used by the computer vision community of what is considered significant.

\paragraph{Does pretraining on larger datasets mitigate variability induced by the choice of seed?}
It certainly reduces the variations due to using different seeds, but it does not mitigate it. On Imagenet, we found a difference between the maximum and the minimum accuracy of around $0.5\%$, which is commonly accepted as significant by the community for this dataset.

Of course, there are many shortcomings to this study as already discussed in \autoref{sec:limiations}. Yet, I would argue that this is a realistic setup for modeling a large set of recent work in computer vision. We almost never see results aggregated over several runs in publications, and given the time pressure that the field has been experiencing lately, I highly doubt that the majority of them took the time to ensure that their reported results where not due to a lucky setup.

As a matter of comparison, there are more than $10^4$ submissions to major computer vision conferences each year. Of course, the submissions are not submitting the very same model, but they nonetheless account for an exploration of the seed landscape comparable in scale of the present study, of which the best ones are more likely to be selected for publication because of the impression it has on the reviewers. For each of these submissions, the researchers are likely to have modified  many times hyper-parameters or even the computational graph through trial and error as is common practice in deep learning. Even if these changes where insignificant in terms of accuracy, they would have contributed to an implicit scan of seeds. Authors may inadvertently be searching for a lucky seed just by tweaking their model. Competing teams on a similar subject with similar methods may unknowingly aggregate the search for lucky seeds.

I am definitely not saying that all recent publications in computer vision are the result of lucky seed optimization. This is clearly not the case, these methods work. However, in the light of this short study, I am inclined to believe that many results are overstated due to implicit seed selection - be it from common experimental practice of trial and error or of the \emph{``evolutionary pressure''} that peer review exerts on them.

It is safe to say that this doubt could easily be lifted in two ways. First, by having a more robust study performed. I am very much interested in having a study scanning $10^4$ to $10^5$ seeds in a large scale setup, with big state-of-the-art models trained from scratch. That would give us some empirical values of what should be considered significant and what is the effect of randomness. Second, the doubt disappears simply by having the community be more rigorous in its experimental setup. I strongly suggest aspiring authors to perform a randomness study by varying seeds - and if possible dataset splits - and reporting average, standard deviation, minimum and maximum scores.


\acks{This work was granted access to the HPC resources of IDRIS under the allocation 2020-AD011011308 made by GENCI.}

\vskip 0.2in
\bibliography{sample}

\end{document}